\documentclass[10pt,twocolumn,letterpaper]{article}

\usepackage{iccv}
\usepackage{times}
\usepackage{epsfig}
\usepackage{graphicx}
\usepackage{amsmath}
\usepackage{amssymb}
\usepackage{booktabs}
\usepackage{soul}
\usepackage[accsupp]{axessibility}

\usepackage[breaklinks=true,bookmarks=false]{hyperref}

\iccvfinalcopy % *** Uncomment this line for the final submission

\def\iccvPaperID{5} % *** Enter the ICCV Paper ID here

\ificcvfinal\fi

\begin{document}

%%%%%%%%% TITLE
\title{Towards Efficient Point Cloud Graph Neural Networks\\Through Architectural Simplification}

\author{Shyam A.~Tailor\thanks{Work completed during Shyam's internship at Arm ML Research Lab. Correspondence to Shyam Tailor and Partha Maji.}\\
University of Cambridge\\
{\tt\small sat62@cam.ac.uk}
\and
{Ren\'{e} de Jong \hspace{0.2cm} Tiago Azevedo \hspace{0.2cm} Matthew Mattina \hspace{0.2cm} Partha Maji}\\
Arm ML Research Lab\\
{\tt\small first.last@arm.com}
}

\maketitle
\ificcvfinal\thispagestyle{empty}\fi

%%%%%%%%% ABSTRACT
\begin{abstract}
In recent years graph neural network (GNN)-based approaches have become a popular strategy for processing point cloud data, regularly achieving state-of-the-art performance on a variety of tasks.
To date, the research community has primarily focused on improving model expressiveness, with secondary thought given to how to design models that can run efficiently on resource constrained mobile devices including smartphones or mixed reality headsets.
In this work we make a step towards improving the efficiency of these models by making the observation that these GNN models are heavily limited by the representational power of their first, feature extracting, layer.
We find that it is possible to radically simplify these models so long as the feature extraction layer is retained with minimal degradation to model performance; further, we discover that it is possible to improve performance overall on ModelNet40 and S3DIS by improving the design of the feature extractor.
Our approach reduces memory consumption by 20$\times$ and latency by up to 9.9$\times$ for graph layers in models such as DGCNN; overall, we achieve speed-ups of up to 4.5$\times$ and peak memory reductions of 72.5\%.
\end{abstract}

%%%%%%%%% BODY TEXT
\section{Introduction}
3D scanning technology is rapidly becoming ubiquitous as the underlying sensors become smaller, cheaper, and more efficient~\cite{velabit}.
These sensors can produce point clouds, which are unordered sets of points in 3D space.
This is in contrast to other common modalities, such as images, where we have a regular grid structure.
Handling this unstructured representation of data is difficult, and a variety of approaches have been proposed.
Early approaches in the literature include projecting the points onto planes, and applying convolutional neural networks (CNNs)~\cite{mvcnn}, or voxelizing the point cloud and applying 3D convolutions~\cite{octnet}.
In recent years, however, a dominant approach has been to construct graphs from the point cloud, and apply graph neural networks (GNNs)~\cite{pointnet++, dgcnn}.
These approaches continue to achieve high performance on tasks including object classification, part segmentation, and semantic segmentation, with many recent works investigating variations in layer design to improve model accuracy~\cite{paconv, pointtransformer}.

Given the rising popularity of 3D scanning sensors in mobile devices such as smartphones~\cite{iphone} or mixed reality headsets~\cite{ar_simulations}, it is natural to investigate how to design resource-efficient models that can run \emph{on-device}.
Unlike some other fields of computer vision, there has been relatively little research into how to design more efficient models, with the majority of works exclusively focusing on how to increase model accuracy, with secondary thought given to considerations such as memory consumption or latency.
This motivates our work: we seek to understand where it is possible to apply simplifications to point cloud architectures.
With a greater understanding of which design choices affect the accuracy of the models, we aim to radically simplify popular existing model architectures, as a first step towards enabling on-device models for point cloud data.

\begin{figure}
    \centering
    \includegraphics[width=\linewidth]{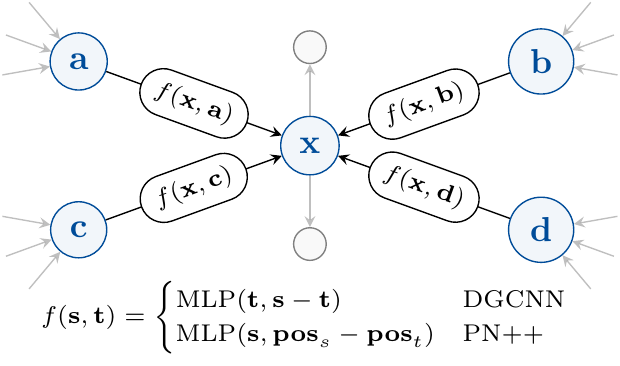}
    \caption{Popular approaches for handling point clouds with GNNs require each message to be explicitly computed and materialized as they are a function of both the source ($\mathbf{s}$) and target ($\mathbf{t}$) nodes.
    This significantly increases memory and latency.
    We illustrate in this paper that many of these operations can be safely removed so long as an effective feature extraction layer is retained.}
    \label{fig:materialization}
\end{figure}

A key challenge with GNNs is their resource demands.
Many state-of-the-art GNNs~\cite{gat, pna}, including point-cloud specific models~\cite{dgcnn, pointnet++} require memory and OPs proportional to the number of edges in the graph ($\mathcal{O}(E)$), as they use complicated anisotropic mechanisms to create the messages that nodes send to each other.
In contrast, many simpler GNN architectures~\cite{gcn, gin} can be implemented using memory and OPs proportional to the number of vertices ($\mathcal{O}(V)$), and it was recently shown by Tailor et al.~\cite{egc} that $\mathcal{O}(V)$ GNN architectures could surpass state-of-the-art models on a variety of tasks.
Given that the number of edges is typically 20-40$\times$ larger than the number of vertices for graphs in point cloud models, this represents a significant opportunity for optimization: the MLP layers constitute most of the latency~\cite{keops}.
This can be seen in Figure \ref{fig:materialization}.
Unfortunately, however, naively applying standard GNN layers to point cloud data yields poor results as they fail to incorporate geometric priors.
In this work, we investigate which components are most critical to point cloud model performance so that we can incorporate optimizations already known in the wider GNN community.

Succinctly, we hypothesise that the most critical layer to model performance is the very first one acting on the raw point cloud data.
We find support for our hypothesis by evaluating it on two datasets, ModelNet40 and S3DIS, with two models, PointNet++~\cite{pointnet++} and DGCNN~\cite{dgcnn}.
Although this hypothesis is relatively simple, it has two useful implications: firstly, if we aim to increase accuracy, it is worth investing more resources into this feature extracting block, and specialising it for the input data.
Secondly, it is possible to simplify the rest of the layers in the model, with little impact on accuracy, but offering large improvements in efficiency.
Together, these observations provide a strong step towards designing more efficient GNN-based models for point cloud data.

In summary, our work makes the following contributions:

\begin{enumerate}
    \item We provide extensive ablation studies on two of the most popular GNN-based architectures to point cloud processing, PointNet++ and DGCNN.
    Our experiments demonstrate the validity of our hypothesis regarding the importance of the feature extraction layer.
    \item We demonstrate how to improve the feature extraction layer by incorporating stronger geometric priors.
    Our approach enables DGCNN to achieve close to state-of-the-art results on ModelNet40, and yields improvements of more than 2 mIoU on S3DIS.
    \item We illustrate that our approach reduces memory consumption by up to 72.5\% in practice.
    We achieve speed-ups of up to 4.2$\times$ on CPU, and 2.5$\times$ on GPU.
\end{enumerate}

\section{Related Work}

\subsection{Architectures for Point Clouds}
Multiple approaches have been proposed for processing point cloud data using neural networks; this section describes the most popular approaches.

\paragraph{Mapping 3D Points to 2D Images}
Multi-view CNN~\cite{mvcnn} approaches involve projecting the point cloud onto 2D planes, and feeding the resulting depth images into a CNN model.
In recent years, these approaches have fallen out of favour, however a recent work has shown that they still offer a strong baseline for point cloud classification~\cite{simpleview}.
Despite this, there are significant weaknesses associated with these methods: they struggle with non-uniform or low point density, and occlusions.

\paragraph{3D Convolutions}
Another approach is to discretize the 3D space into voxels, and apply 3D convolutions~\cite{octnet, voxnet, 3dmfv, vvnet}.
The primary weakness with this approach is the inevitable information loss when voxelizing the point cloud; this can be mitigated by using a finer discretization, but this results in greater memory and computational costs.

\paragraph{Point Convolutions}
PointCNN~\cite{pointcnn} proposed learning transformations that enable the convolution to be done in a different, regular, space; however, this approach does not retain permutation-invariance, which restricts representational power.
Other approaches~\cite{spiderconv, pointconv} define \textit{continuous} kernels; implementation requires explicit prediction of the kernel weights, resulting in high computational cost.
KPConv~\cite{kpconv} proposes to simplify these approaches by restricting the convolution to be defined by a set of anchor points, reducing the computational complexity and improving accuracy.

\paragraph{Graph Neural Networks}
PointNet~\cite{pointnet} proposed directly operating on the input points using shared MLPs, and using global pooling operations to share information across the point cloud.
This approach is highly efficient, but does not allow for local information to be aggregated, restricting the expressiveness of the architecture.
Follow-up works, such as PointNet++~\cite{pointnet++} and DGCNN~\cite{dgcnn} built upon PointNet by explicitly constructing graphs upon which local aggregation is performed, thereby boosting accuracy at the cost of significant increases to memory consumption and runtime.
These models can be interpreted using the message passing paradigm that is commonly used for describing graph neural networks (GNNs)~\cite{gilmer}.
In this paradigm, a source node sends a \textit{message} along a directed edge to a target node; each node \textit{aggregates} the messages it has received using a permutation-invariant function, such as max or mean, and uses the aggregated result to update its representation.
For PointNet++ and DGCNN we have the following rules for updating node representations:

\begin{equation}
    \mathbf{x}_i^{l + 1} = \max_{j \in \mathcal{N}(i)} \text{MLP}(\mathbf{x}_j^l \; || \; \mathbf{p}_j - \mathbf{p}_i) \tag{PointNet++}
\end{equation}
\begin{equation}
    \mathbf{x}_i^{l + 1} = \max_{j \in \mathcal{N}(i)} \text{MLP}(\mathbf{x}_i^l \; || \; \mathbf{x}_j^l - \mathbf{x}_i^l) \tag{DGCNN},
\end{equation}

where $\mathbf{x}_i^l$ is the representation of point $i$ at layer $l$, $\mathbf{p}_i$ represents the 3D position of point $i$, and $\mathcal{N}(i)$ is the set of neighbors of point $i$ in the constructed graph, which is found using kNN for DGCNN and radius queries for PointNet++.
In the first layer, DGCNN represents $\mathbf{x}_i$ as the point features (if any) concatenated with the point position.

In this work we focus on PointNet++ and DGCNN, which are among the most popular architectures, and remain strong baselines in many cases~\cite{simpleview}.
However, the approaches and observations described in this work are agnostic to architecture, and are transferable to more recent approaches achieving state-of-the-art performance~\cite{paconv, pointtransformer}.
It is worth noting that our conclusions will also be useful to point convolution approaches, which also operate on graphs and can be cast as a sub-group of the GNN approaches.

\subsection{Optimizing Graph Neural Networks}
There is relatively little work on optimizing GNN inference.
Quantization~\cite{szebook} is one approach to reduce latency and memory consumption by reducing the bitwidth of the weights and activations.
In practice this is not straightforward for GNNs, with specialist training procedures being required~\cite{degreequant}.
Another avenue of research is architecture design: Zhao et al.~\cite{zhao_PDNAS} studied how to arrange GNN layers to maximise accuracy given certain constraints.
Tailor et al.~\cite{egc} explicitly studied efficient architecture design for GNNs, and proposed an architecture that could achieve state-of-the-art performance while using memory and OPs proportional to the number of vertices in the graph---not the number of edges as is common for many GNN architectures~\cite{gnn_survey}, including PointNet++ and DGCNN.

\paragraph{Point Cloud Model Optimization}
Li et al.~\cite{li_efficientdgcnn} demonstrated that it is possible to significantly reduce the OPs used by DGCNN by decomposing the MLP operation into two separate MLPs: $\text{MLP}_1(\mathbf{x}_i^l) - \text{MLP}_2(\mathbf{x}_j^l)$.
This observation is effective at reducing OPs, but does not reduce memory consumption at inference time, and this work did not investigate how this approach generalized to other architectures.
Another recent work has investigated accelerating the grouping operations these architectures use to construct the graph: KeOps~\cite{keops} achieves a 10-100$\times$ speed-up in practice for these operations.
In practice, the latency bottleneck in these models is now the MLP layers and aggregation operations~\cite{keops}, which are the focus of this work.
Accelerating the MLP layers and aggregations is possible with quantization, which has been studied for point cloud networks by \cite{bipointnet}.
PosPool~\cite{pospool} demonstrated that it is possible to replace the sophisticated aggregation methods proposed by many point cloud works with a simpler method; however, to achieve this, they relied upon a deep residual network, which has significantly higher latency due to the number of aggregation steps.
Finally, RandLa-Net~\cite{randlanet} demonstrated that by encoding individual points in the point cloud with details of their neighborhood, it was possible to use a random downsampling, rather than the slower (but more uniform) farthest point sampling technique, to achieve high accuracy when segmenting large point clouds.
\section{Why Are Point Cloud GNNs Intolerant of Optimizations?}
Although works such as EGC~\cite{egc} have shown great promise at optimizing GNNs for non-point cloud tasks, our early experiments indicated that they do not achieve strong performance on point cloud tasks.
Indeed, we have observed a proliferation of architectures designed for point clouds exclusively, a trend that has not occurred to nearly the same extent for other data modalities GNNs have been applied to.
We ask: \textbf{why do we need point cloud-specific architectures?}

\subsection{Our Hypothesis: Information Bottleneck at the First Layer}
\label{sec:hypothesis}
We hypothesise that point cloud architectures are severely bottlenecked by the information extracted at the first layer, where the primary input is the point positions.
Following this hypothesis, we believe that the reason point cloud-specific architectures achieve better results is because they incorporate geometric priors into this first layer.
Critically, we argue that these geometric priors lose relevance deeper into the network when reasoning about the representations geometrically is less well defined---and are arguably even harmful given their lack of success on other data modalities, and potential to induce overfitting.
There are two important corollaries of this hypothesis:

\begin{enumerate}
    \item To improve accuracy it is best to spend resources improving the feature extracting first layer.
    \item After the first layer, it is possible to radically simplify the architecture to reduce latency and memory consumption.
\end{enumerate}

\subsection{Experimental Verification}
\begin{table}[]
\resizebox{\linewidth}{!}{
\centering
\footnotesize
\begin{tabular}{@{}ccccccc@{}}
\toprule
\multicolumn{1}{l}{} & \multicolumn{1}{l}{} & \multicolumn{2}{c}{\textbf{ModelNet40}}          & \multicolumn{3}{c}{\textbf{S3DIS Area 5}}                                  \\ \cmidrule(l){3-7} 
\textbf{Model}       & \textbf{Experiment}  & \textbf{OA} & \multicolumn{1}{c|}{\textbf{mAcc}} & \multicolumn{1}{c}{\textbf{mIoU}} & \multicolumn{1}{c}{\textbf{OA}}  & \multicolumn{1}{c}{\textbf{mAcc}} \\ \midrule
PN++ SSG             & Baseline                  & 92.8       & 89.5                              &     55.2                              &  84.6   &       64.0                     \\
PN++ MSG             & Baseline                  & 92.5       & 89.4                              &   -                                &    -    &     -                    \\
DGCNN                & Baseline                  & 92.9       & 88.9                              &    53.1                  &  85.3        &       59.8               \\ \midrule
PN++ SSG             & (1)                  & 90.5        & 86.8                               &    54.0                               &     84.1         &     62.2              \\
PN++ MSG             & (1)                  & 91.3        & 87.8                               &    -                               &      -        &          -         \\
DGCNN                & (1)                  & 91.7        & 88.9                               &      49.2                             &     83.3          &        57.6          \\ \midrule
PN++ SSG             & (2)                  & 91.2       & 88.7                               &       54.3                            &      84.6          &        61.8         \\
PN++ MSG             & (2)                  & 90.9       & 87.7                               &      -                             &       -        & -                 \\
DGCNN                & (2)                  & 92.8        & 89.3                               &       51.6                            &     83.9          &            57.6      \\ \midrule
PN++ SSG             & (3)                  & 92.9       & 89.3                              &            54.3                       &        84.5         &          62.3      \\
PN++ MSG             & (3)                  & 92.6       & 88.6                              &     -                              &         -       &             -    \\
DGCNN                & (3)                  & 92.8       & 90.4                              &           54.0                        &       85.2          &            62.5    \\ \bottomrule
\end{tabular}
}
\normalsize
\vspace{0mm} % i have no idea why this is necessary...
\caption{Results of inserting simplified blocks into PointNet++ and DGCNN.
SSG and MSG refer to single and multi-scale grouping PointNet++ variants respectively~\cite{pointnet++}.
Experiment (1) refers to completely replacing all blocks with simplified versions; (2) refers to replacing only the first block; (3) refers to replacing all blocks \textit{except the first one}.
Retaining the first block only yields results comparable, if not better, than the baseline.}
\label{tab:ablate_blocks}
\end{table}

We now provide experimental evidence supporting our hypothesis by evaluating DGCNN and PointNet++ on ModelNet40~\cite{mn40} and S3DIS Area 5~\cite{s3dis}.
As pointed out by Goyal et al.~\cite{simpleview}, precise experimental setup can have a large effect on the end results obtained when working with point cloud data.
We use a similar protocol to ~\cite{paconv}; one change we make is to sample points for ModelNet40 randomly, rather than using farthest point sampling, to make the task more difficult.
To facilitate the interpretation of the results, we run the baselines with our implementations to ensure our results are self-consistent.

Many works in the point cloud literature have argued that ``centralization'' operations, in which node representations are centralized to their local neighborhood, are vital to obtaining strong performance~\cite{pointnet++, dgcnn, pospool}; these can be witnessed in the update rules for both PointNet++ and DGCNN.
However, performing these centralization operations has significant memory and computational overheads, as seen in fig.~\ref{fig:materialization}.
Therefore, we devote this section to assessing when exactly centralization is required, enabling us to evaluate our hypothesis.
In particular, we consider simplified update rules for both PointNet++ and DGCNN:

\begin{equation}
    \mathbf{x}_i^{l + 1} = \max_{j \in \mathcal{N}(i)} \text{MLP}(\mathbf{x}_j^l \; || \; \mathbf{p}_j) \tag{PointNet++-Simple}
\end{equation}
\begin{equation}
    \mathbf{x}_i^{l + 1} = \max_{j \in \mathcal{N}(i)} \text{MLP}(\mathbf{x}_j^l) \tag{DGCNN-Simple}
\end{equation}

The reader should also note that both these simplified operators significantly reduce the memory consumption and OP count.
These simplified operators can be implemented using $\mathcal{O}(V)$ memory and OPs, and the propagation step can be implemented using matrix multiplication-style algorithms, which are more practical to accelerate~\cite{egc}.
In contrast, the baseline operators use $\mathcal{O}(E)$ memory and OPs.
Therefore, if our hypothesis holds, we can achieve savings of approximately 20-40$\times$ in both quantities compared to the baselines in the graph layers.

We present experiments on ModelNet40 and S3DIS Area 5 in Table~\ref{tab:ablate_blocks}.
We focus on ablating the contribution of the baseline geometrically-inspired update rules, by evaluating their contribution at the first layer, and beyond the first layer.
Our experiments clearly demonstrate that using a simplified block after the first layer does not significantly reduce accuracy, and in some cases, actually improves it.
In contrast, when a simplified block is used as the first layer, there is a noticeable drop in accuracy.
It is also not the case that a sophisticated backbone can recover the information lost in the first layer: we observe only changing the first layer still results in a noticeable accuracy drop.
This trend is especially clear for DGCNN; unlike PointNet++, there is no additional geometric information injected at later layers of the model, in the form of point positions, hence we expect that it is more reliant on the feature extractor.
We expect the performance of the simplified architectures---although superior to the baselines in many cases---can be improved further with better hyperparameters, which have been tuned for the baselines exclusively.

These results support our hypothesis, and motivate the rest of this work with a natural follow-up question: \textbf{can we improve the feature extractor to improve accuracy, while retaining low overall resource usage?}

\section{Designing Accurate and Efficient Feature Extractors}
In this section we investigate how to construct better feature extractors from our point cloud, and show that incorporating basic geometric information into the first feature extracting layer, and retaining a simplified backbone, can enable us to improve model accuracy.
We use DGCNN as a motivating case study, however our experiments can equally be applied to other backbones; we find that our modifications enable us to obtain close to state-of-the-art performance on ModelNet40\footnote{The reader should note that our experiments are run using non-uniform point densities, making our task more difficult than most results in the literature}, and increases of more than 2 mIOU on S3DIS relative to an unmodified DGCNN model.
This confirms corollary (1) from our hypothesis described in section \ref{sec:hypothesis}, where we stated that improving feature extractors can enable us to achieve better accuracy overall.

\subsection{Finding Useful Geometric Features}
\label{sec:feat_extractor}
It is well known that adding information such as normal vectors or classical point cloud feature histograms to the input point cloud will boost accuracy; we are making a related observation that simply incorporating stronger geometric priors into the first, feature-extracting, layer of the architecture can serve as a cheap and effective way to boost accuracy~\cite{pointnet, srivastava}.
These geometric priors do not need to be repeated throughout the model, affording us the use of efficient graph layers elsewhere.

\begin{table}[]
\centering
\footnotesize
\begin{tabular}{@{}cccc|cc@{}}
\toprule
\textbf{Source Pos} & \textbf{Target Pos} & \textbf{Rel Pos} & \textbf{Distance} & \textbf{OA} & \textbf{mAcc} \\ \midrule
        \checkmark            &           \checkmark          &          \checkmark        &         \checkmark          &      93.2       &       90.2        \\ \midrule
        &           \checkmark          &          \checkmark        &         \checkmark          &       \textbf{93.4}      &     90.4          \\
        \checkmark            &           &          \checkmark        &         \checkmark          &     \textbf{93.4}        &   \textbf{90.7}            \\
        \checkmark            &           \checkmark          &          &         \checkmark          &   92.8          &   90.0            \\
        \checkmark            &           \checkmark          &          \checkmark        &         &     92.9        &     89.9          \\ \midrule
        &     \checkmark      &          \checkmark        &         &        92.8     &       90.4        \\ \bottomrule
\end{tabular}
\normalsize
\vspace{2mm}
\caption{
Results of ablation study on feature extractor on ModelNet40.
We assess removing each individual component of the feature extractor to assess their contribution to end performance.
The bottom row is DGCNN's standard block, which can be interpreted as implementing a subset of the operations included in our feature extractor.
We observe that adding features such as distance can noticeably increase accuracy to close to state-of-the-art.
}
\label{tab:extractor}
\end{table}

To investigate the contribution of different geometric priors at the first layer, we consider the use of a new update rule consisting of 4 features applied to ModelNet40.
The geometric features we identify are: (1) Source position $\mathbf{p}_j$; (2) Target position $\mathbf{p}_i$; (3) Relative position $\mathbf{p}_j - \mathbf{p}_i$, and (4) Euclidean distance $\text{dist}(\mathbf{p}_j, \mathbf{p}_i)$.
We form messages from the source to the target node by concatenating the chosen features together and applying an MLP; the reader should note that this update rule simply extends the default DGCNN update rule at the first layer with additional features.
All layers other than the first use the simplified DGCNN layers described in the previous section.

The results of our ablation study on the proposed feature extractor are shown in Table \ref{tab:extractor}.
We observe in the top row that including all 4 features yields an increase in overall accuracy; however, removing either source or target positions from the update rule can boost accuracy further.
In contrast, removing relative positional information or distance information noticeably reduces the accuracy.
We believe that the increase in accuracy due to removing source or target position is due to reduced overfitting; the reader should note that providing all three positional components is redundant, as the third component can be directly computed from the other two provided.
However, given that the removal of relative positional information noticeably impacts accuracy, we believe that faithful reconstruction of relative position from source and target positions cannot be relied upon without careful or lucky weight initialization.
As the relative position is more critical to model performance than either the raw source or target positions, it is better to explicitly compute it before feeding into the model.
Similarly, distance cannot be computed without multiple layers (as it is a non-linear transform), increasing the required OPs and memory, and making training more difficult~\cite{pospool}.

\paragraph{Improving Performance on S3DIS}
With our improved understanding of how to handle positional data, we proceed to demonstrate that these gains translate to the more difficult S3DIS dataset.
Unlike ModelNet40, where the input data consists exclusively of raw point positions, S3DIS introduces RGB color information for each point.

Building upon our conclusions in the previous section, we propose the following rule for a feature extractor on S3DIS:
\begin{equation}
\scriptstyle
    \mathbf{x}_i^{l + 1} = \max_{j \in \mathcal{N}(i)} \text{MLP}(\mathbf{x}_j^l \, || \, \mathbf{x}_j^l - \mathbf{x}_i^l \, || \, \mathbf{p}_j^l - \mathbf{p}_i^l \, || \, \mathbf{p}_j^l \, || \, \text{dist}(\mathbf{p}_j, \mathbf{p}_i)).
    \label{eq:s3dis_extractor}
\end{equation}

\begin{table}[]
\centering
\begin{tabular}{@{}cccc@{}}
\toprule
\textbf{Aggregators} & \textbf{mIoU} & \textbf{OA} & \textbf{mAcc} \\ \midrule
Baseline (tab.~\ref{tab:ablate_blocks})             & 53.2         & 85.3       & 62.5         \\
max                  & {55.2}          & 85.4        & {63.2}          \\
max + min            & 53.6         & {86.0}       & 60.8         \\
max + mean           & 55.3         & 84.9       & {63.2}         \\ \bottomrule
\end{tabular}
\vspace{2mm}
\caption{Performance of the proposed feature extractor on S3DIS.
We observe that our feature extractor improves performance over the baseline DGCNN model by 2 mIoU.
Adding additional aggregators, although effective on other GNN datasets~\cite{pna}, does not yield immediate improvements for point cloud data.}
\label{tab:s3dis_extractor}
\end{table}

In the GNN literature it is well known that incorporating multiple aggregators can boost performance~\cite{pna}.
To assess whether this observation also applies to point cloud data, we also evaluate replacing the max aggregator with combinations of multiple aggregators using the approach proposed by \cite{pna}; although the $\max$ aggregator provides strong performance when operating on geometric data, it may be sub-optimal when applied to color information.
We investigate adding additional min and mean aggregators to assess whether they aid feature extraction.

The results obtained on S3DIS Area 5 are presented in Table \ref{tab:s3dis_extractor}.
Our proposed feature extractor improves performance over the baseline DGCNN model by 2 mIoU, with similar computational overhead.
We observe that adding the min aggregator does improve the architecture's ability to discriminate local geometry, however the increased representational power leads to overfitting rather than generalization; in contrast, we find that the addition of the mean aggregator offers negligible benefit to accuracy overall.

\paragraph{Summary}
We have demonstrated the value of designing specialized feature extractors when working with point cloud data.
Despite using simplified backbones, we have shown that the addition of these feature extractors can noticeably boost accuracy beyond the baseline architecture.
Our conclusions provide new insight for future works seeking to design efficient, but still expressive, architectures operating on point clouds.

\section{Further Studies}
In this section we present a collection of studies to expand our conclusions from the previous sections.
We first show that it is possible to get near state-of-the-art performance on ModelNet40 using Transformers~\cite{transformer} if a good feature representation is provided.
We also present studies on how to design simplified feature extraction layers that trade accuracy for reduced resource consumption.
Finally, we provide an in-depth analysis of our approach's robustness and resource consumption.

\subsection{Feature Extraction for Transformers}
In recent years we have witnessed Transformers achieving state-of-the-art performance across tasks including natural language processing~\cite{transformer}, speech recognition~\cite{conformer}, and image classification~\cite{vit}.
However, we have yet to see unmodified Transformer models achieving this level of performance on point cloud tasks.
Fitting with the hypothesis underpinning this work, we believe that Transformers can only achieve good performance if appropriate features are provided as input to the Transformer layers: raw point positions alone are insufficient.
One approach to this issue is to modify the Transformer layers, such that they have only superficial resemblance to the original architecture~\cite{pointtransformer}; however, adopting this approach precludes the usage of advances in Transformer models in the literature---especially with regard to improvements in efficiency.
We adopt an alternative approach of using our feature extraction layer, and feeding the resulting features into the Transformer layers.

We use an efficient Transformer variant, the Performer~\cite{performer}, to reduce the memory consumption to linear in the number of points.
The model used 4 layers with a dimension of 256 and 4 heads; the feed forward layer dimension was 1024.
These parameters were not extensively optimized; we expect with further investigation our results can be improved.
The primary aim of this study is to demonstrate the importance of choices in feature extraction, even with little to no fine-tuning.

\begin{table*}[ht]
\centering
\footnotesize
\begin{tabular}{@{}cccccccc@{}}
\toprule
                 &                                 & \multicolumn{4}{c}{\textbf{Feature Extractor}}                                                 &             &               \\ \cmidrule(lr){3-6}
\textbf{Raw Position} & \textbf{RPPE}~\cite{randlanet}                   & \textbf{Source Position} & \textbf{Target Position} & \textbf{Rel Position} & \textbf{Distance}               & \textbf{OA} & \textbf{mAcc} \\ \midrule
\checkmark       & \multicolumn{1}{c|}{}           &                     &                     &                  & \multicolumn{1}{c|}{}           &  90.4           & 86.3              \\
                 & \multicolumn{1}{c|}{\checkmark} &                     &                     &                  & \multicolumn{1}{c|}{}           &  91.3          & 87.0             \\ \midrule
                 & \multicolumn{1}{c|}{}           & \checkmark          & \checkmark          & \checkmark       & \multicolumn{1}{c|}{\checkmark} &  93.1           & 88.8           \\
                 & \multicolumn{1}{c|}{}           &                     & \checkmark          & \checkmark       & \multicolumn{1}{c|}{\checkmark} &  92.9           & 89.5           \\
                 & \multicolumn{1}{c|}{}           & \checkmark          &                     & \checkmark       & \multicolumn{1}{c|}{\checkmark} &  92.8           & 88.8           \\
                 & \multicolumn{1}{c|}{}           & \checkmark          & \checkmark          &                  & \multicolumn{1}{c|}{\checkmark} &  \textbf{93.2}           & \textbf{89.9}           \\
                 & \multicolumn{1}{c|}{}           & \checkmark          & \checkmark          & \checkmark       & \multicolumn{1}{c|}{}           &  92.9           & 89.1           \\ \midrule
                 & \multicolumn{1}{c|}{}           &                     & \checkmark          & \checkmark       & \multicolumn{1}{c|}{}           &  92.9           & \textbf{89.9}               \\ \bottomrule
\end{tabular}
\vspace{2mm}
\caption{Results when combining different feature extractors with a Transformer on ModelNet40; the bottom row corresponds to using a DGCNN-style block for feature extraction.
We observe that accuracy improves as the Transformer is combined with more sophisticated feature extractors.
Unlike Table \ref{tab:extractor}, we observe that it is better to include source and target positions rather than the relative position.
We conjecture that this is due to the query-based attention mechanism used in Transformers.}
\label{tab:transformer}
\end{table*}

Our study considers a variety of techniques to feed the points into the Transformer; we consider feeding the raw point positions in after applying a single fully-connected layer (similar to ViT~\cite{vit}); relative point position encoding (RPPE) from~\cite{randlanet}, in which each point is encoded with the absolute and relative positions of its nearest neighbors; DGCNN's feature extractor, and our proposed feature extractor from section \ref{sec:feat_extractor}.
The results are presented in Table \ref{tab:transformer}.
As expected, feeding the raw positional features into the model attains poor performance; RPPE improves upon this by encoding local geometric information into each points representation, but still achieves relatively poor performance.
In contrast, using the DGCNN-based feature extractor (bottom row) yields performance close to that obtainable with a unmodified DGCNN model.
As before, incorporating the full set of features in the feature extractor improves performance further.
In contrast to our conclusions in section \ref{sec:feat_extractor}, it is observed that removing relative point position is most effective for reducing overfitting.
One possible explanation for this observation is that the attention mechanism is more effective when each point's representation also encodes absolute positional information about neighbors; this also explains the effectiveness of RPPE over the raw point positions.

The results in this section provide further support for our hypothesis regarding the importance of feature extraction.
One observation we make is that the performance achieved by the Transformer models when combined with the feature extractor layers corresponds closely to the results we achieved in section \ref{sec:feat_extractor}.
This is further evidence that the feature extraction layer limits the overall model performance.

\subsection{Designing Linear-Memory Feature Extractors}
\label{sec:lin_extractor}
The feature extractors we have described in this paper require memory and computation proportional to the number of edges in the constructed graph, as we explicitly construct the message for each edge in the graph as a function of both source and target nodes.
While this is acceptable in many cases, it may represent a bottleneck for resource constrained devices; it is also problematic due to the increased data movement which is a significant contributor to power consumption~\cite{horowitz}.
Another challenge is that message propagation and aggregation must be implemented using index selection operations which are not typically optimized on mobile GPUs or NPUs.
As explained earlier, simplified GNN layers can implement these two steps using matrix multiplication-style approaches which are well supported in hardware.
These issues motivate us to investigate if we can design feature extractor layers that attain high performance while using $\mathcal{O}(V)$ memory.

This work has already emphasized the importance of the centralization operations in the feature extraction layer.
However, it is not possible to use these operations and continue to use matrix multiplication for message propagation around the graph: our procedure to create messages must be a function of the source node only.
To this end, we propose the following update rule for a ModelNet40 feature extractor to imitate the centralization operations:

\begin{equation}
    \mathbf{x}_i^{l + 1} = \left(\max_{j \in \mathcal{N}(i)} \text{MLP}_1(\mathbf{p}_j)\right) - \text{MLP}_2(\mathbf{p}_i).
    \label{eq:lin_extractor}
\end{equation}

Each point is passed through an MLP, and the resulting representation is subtracted from the result aggregated from the local neighborhood.
We consider sharing the weights across MLPs, and learning separate MLPs.
In principle using separate MLPs should increase model capacity, at the cost of less stable optimization, increased runtime, and potential overfitting.

\begin{table}[h]
\centering
\footnotesize
\begin{tabular}{@{}cccccc@{}}
\toprule
\textbf{Experiment}        & \multicolumn{2}{c}{\textbf{ModelNet40}}          & \multicolumn{3}{c}{\textbf{S3DIS Area 5}}   \\ \cmidrule(l){2-6} 
                           & \textbf{OA} & \multicolumn{1}{c|}{\textbf{mAcc}} & \textbf{mIoU} & \textbf{OA} & \textbf{mAcc} \\ \midrule
\textbf{No centralization} & 91.7       & 88.9                              & 49.2         & 83.3       & 57.6         \\
\textbf{Shared MLPs}       & 92.3       & 89.2                              & \textbf{53.5}         & \textbf{84.3}       & \textbf{61.4}         \\
\textbf{Separate MLPs}     & \textbf{93.2}       & \textbf{90.6}                              & 50.4         & 84.1       & 57.4         \\ \bottomrule
\end{tabular}
\vspace{2mm}
\caption{Results when using our resource-efficient feature extraction layer.
We observe that adding centralization increases performance on both datasets.
Sharing the MLP weights achieves better performance on S3DIS, in contrast to ModelNet40, where adding a separate MLP improves performance beyond the DGCNN baseline.
This is due to S3DIS backbone using deeper MLPs in the feature extractor, which is known to induce overfitting and make the optimization less stable~\cite{pospool}}
\label{tab:lin_extractor}
\end{table}

The results when using this feature extractor are shown in Table \ref{tab:lin_extractor}.
We provide results using our proposed rule and where no centralization is applied at all.
As expected, not using centralization causes noticeable degradation in accuracy.
Impressively, our proposed approach surpasses the performance of the full baseline DGCNN model on both ModelNet40 and S3DIS reported in Table \ref{tab:ablate_blocks}, although it does not reach the performance of the full feature extraction approach proposed in section \ref{sec:feat_extractor}.
This is unsurprising given the reduction in model complexity.

Our results indicate that it is better to use separate weights for the MLPs on ModelNet40, and share them on S3DIS.
Although this observation is initially counter-intuitive, it is consistent with observations made by Liu et al.~\cite{pospool}, where they find that using deeper MLPs in graph layers can impede optimization.
The feature extraction layer for the ModelNet backbone uses MLPs of depth 1, in contrast to S3DIS which uses MLPs of depth 2; our ablated architectures copy the depths faithfully.
By sharing the weights, we reduce the potential for overfitting and improve the stability of the optimization, therefore yielding better generalization---even if the theoretical model capacity is reduced.

\subsection{Model Robustness and Resource Usage}
% Model performance metrics are not the only objectives to optimize for.
In the real world, we are also interested in ensuring that our model tolerates imperfect inputs, such as missing points, or high noise.
We also care about model latency, memory consumption, and hardware support.

\paragraph{Model Robustness}
\begin{figure}
    \centering
    \includegraphics[width=\linewidth]{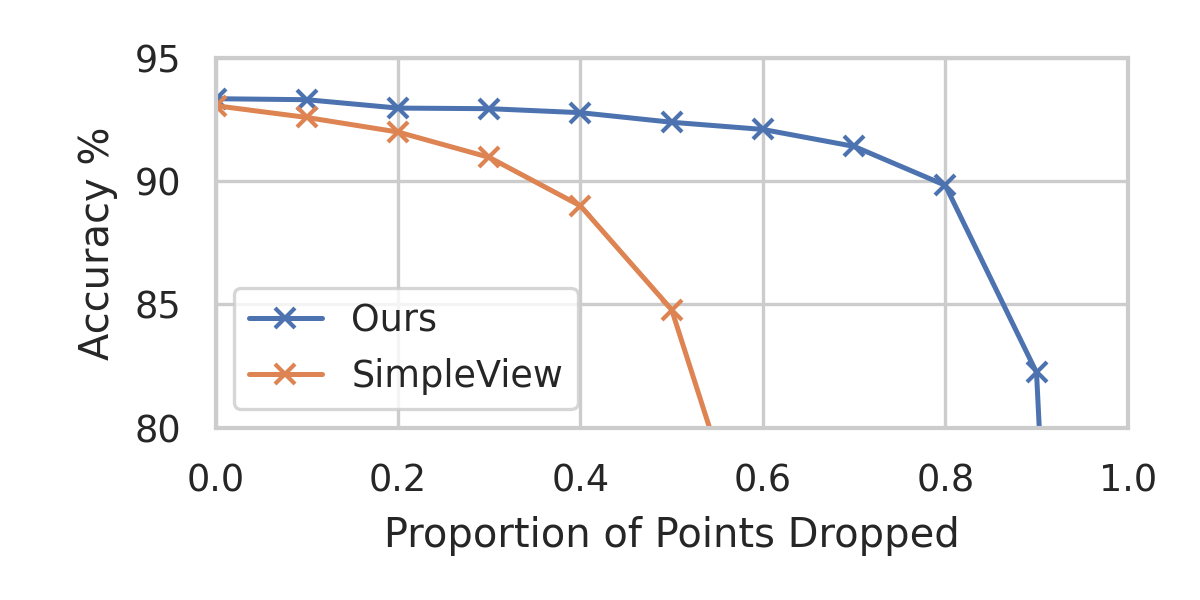}
    \caption{Comparison of our approach with SimpleView, a multi-view CNN, on ModelNet40.
    Even when 90\% of the points have been dropped our approach still obtains accuracy above 80\%.
    Although multi-view methods are easier to optimize, they have significant downsides with regard to robustness.}
    \label{fig:robustness}
\end{figure}
As demonstrated by Goyal et al.~\cite{simpleview}, multi-view CNNs still represent a strong baseline for point cloud tasks.
Given that CNNs have been the target of much research investigating how to make them more efficient, and are well supported in hardware, it could be argued that it makes more sense to optimize this class of models instead of GNNs.
In practice, a strong argument against this is that multi-view models are not as robust to variations in point density, or missing points.
To illustrate this visually we plot accuracy on ModelNet40 as the proportion of points dropped increases in fig.~\ref{fig:robustness}.
We include our full model (section \ref{sec:feat_extractor}) alongside the SimpleView multi-view CNN model~\cite{simpleview}.
It can be observed that our models retain high accuracy, even as 95\% of points are dropped; in contrast, SimpleView's accuracy rapidly degrades.
We found that deviating from the recommended hyperparameters for SimpleView could enable more robust models to be trained, but with a steep 2\% decrease in overall accuracy when evaluating on unmodified point clouds.

\paragraph{Inference Latency}
\begin{table}[]
\resizebox{\linewidth}{!}{
\centering
\footnotesize
\begin{tabular}{@{}ccccc@{}}
\toprule
\textbf{Model} & \multicolumn{2}{c}{\textbf{ModelNet40}}          & \multicolumn{2}{c}{\textbf{S3DIS}} \\ \cmidrule(l){2-5} 
               & \textbf{CPU} & \multicolumn{1}{c|}{\textbf{GPU}} & \textbf{CPU}     & \textbf{GPU}    \\ \midrule
DGCNN          &  96.3 $\pm$ 2.3            &   3.00 $\pm$ 0.04                               &     260.0 $\pm$ 1.9            & 12.31 $\pm$ 0.36                 \\
Full Extractor (sec. \ref{sec:feat_extractor})          &   25.7 $\pm$ 1.1          &    1.20 $\pm$ 0.05                               &    158.7 $\pm$ 1.8             &   9.13 $\pm$ 0.06              \\
Full Extractor (sec. \ref{sec:feat_extractor}) + SpMM    &  23.0 $\pm$ 0.9            & 1.19 $\pm$ 0.06                                 &   154.1 $\pm$ 1.9               & 9.04 $\pm$ 0.13               \\
Efficient Extractor (sec. \ref{sec:lin_extractor}) + SpMM    & 17.7 $\pm$ 0.6            &  1.09 $\pm$ 0.08                                 &   101.3 $\pm$ 1.5               & 7.71 $\pm$ 0.20                 \\ \bottomrule
\end{tabular}
}
\vspace{0mm}
\caption{Inference times for our models compared to the baseline DGCNN.
We observe that our approaches provide especially large speed-ups when using the ModelNet40 backbone, and when running on CPU.
Using SpMM provides a small reduction to latency on CPU, in addition to its main benefit of reducing peak memory consumption.
Incorporating our efficient feature extractor further reduces model latency.
}
\label{table:latency}
\end{table}

In Table \ref{table:latency} we provide inference latency of the baseline DGCNN model, alongside our models which use simplified backbones.
We ran our models using PyTorch 1.9; the CPU was an Intel i9-7900x and the GPU was an Nvidia RTX 2080.
On S3DIS we are timing the latency to process a single batch of 4096 points.
We observe that our approach yields speed-ups of 4.2$\times$ and 1.7$\times$ respectively for ModelNet40 and S3DIS on CPU; on GPU we see speed-ups of 2.5$\times$ and 1.4$\times$ respectively.
Further speed-ups are obtained when the feature extraction layer uses the resource efficient approaches proposed in section \ref{sec:lin_extractor}.
Although we have achieved significant speed-ups (especially on CPU), we cannot reduce latency further without optimizing other bottlenecks due to Amdahl's Law; the DGCNN backbone we are modifying has very wide fully connected layers after the graph layers, which contribute a large fraction of the latency.
We also note that we have not focused on accelerating the sparse propagation operations in the graph layer: these represent a noticeable bottleneck, especially on GPUs.
For a single layer with input and output dimension of 128, we observe speed-ups of 9.9$\times$ on CPU and 3.5$\times$ on GPU.
To improve end-to-end latency further requires rethinking the model backbone design, which lies beyond the scope of this work.

\paragraph{Memory Consumption}
For DGCNN models we record peak memory consumption to be 69.7MB on ModelNet40 and 129.6MB on S3DIS.
By comparison, our models achieve 19.2MB and 100.3MB peak consumption respectively.
Although our simplified layers do achieve a measured 20$\times$ improvement in memory consumption, once again we do not realise the full benefit of our approach due to the wide fully connected layers that are prevalent in DGCNN's architecture design.

\section{The Road Ahead}
% This work provides useful insight into what aspects of point cloud GNN architectures are vital for increasing accuracy, and demonstrates that simplifications can be applied.
Our insight is only one detail with regards to optimizing these models for deployment on resource-constrained devices.
% There are several other aspects of architecture design that may enable further reductions to latency and memory consumption.
Of critical importance is rethinking the backbone architecture design: while our approach has improved the efficiency of these models, we can achieve stronger results by re-designing the backbone with reference to the target hardware implementation.

\paragraph{Going Deeper}
Most works utilize backbones that are shallow; for example, DGCNN and PointNet++ backbones use 3-4 graph layers.
As shown by PosPool~\cite{pospool}, going deeper---$\geq$10 graph layers---in combination with a simple layer design can achieve state-of-the-art performance on point cloud tasks.
While this strategy is effective for boosting accuracy and retaining low peak memory consumption, it will inevitably increase latency due to the expense of the aggregation operations in GNNs.
In contrast, we wish to make the network shallow without sacrificing accuracy.
% This line of research has been explored in the GNN literature~\cite{sgc_wu}, but to the best of our knowledge has not been investigated for point clouds.

\paragraph{Increasing Arithmetic Intensity}
% Accelerators for mobile devices will typically have memory bandwidth of approximately 8GB/s to DRAM.
In order to achieve peak accelerator utilization it is vital to achieve high arithmetic intensity: performing many OPs/byte read from memory.
Our initial calculations indicate that many elements of the DGCNN architecture achieve arithmetic intensity of just 64 OPs/byte, far below what is required to saturate the a mobile accelerator with limited memory bandwidth.
% These observations do not manifest as clearly on data center GPUs, which have significantly more memory bandwidth than mobile NPUs.
Increasing the memory bandwidth is often not possible, hence we must investigate approaches to increase the OPs/byte.
One strategy would be quantization; beyond this, however, requires a novel re-design of our neural network architecture.

\subsection{Conclusion}
We have investigated PointNet++ and DGCNN, and shown that it is possible to simplify them so long as we retain a sophisticated feature extraction layer.
Following our observation, we have shown that by improving the feature extraction layer, we can improve accuracy on ModelNet40 and S3DIS overall despite using resource-efficient backbones.
We discuss the improvements to memory consumption and latency our approaches provide; we observe up to 4.5$\times$ speed-ups on CPUs.
We also show that our observation regarding feature extraction also applies to Transformers processing point clouds.
Finally, we discuss the next steps required to improve performance of point cloud models on mobile devices.

{\small
\bibliographystyle{ieee_fullname}
\bibliography{egbib}
}

\end{document}